# Fundamental principles of cortical computation: unsupervised learning with prediction, compression and feedback

Micah Richert*
richert@braincorporation.com

Dimitry Fisher
fisher@braincorporation.com

Filip Piekniewski
piekniewski@braincoproration.com

Eugene M. Izhikevich
izhikevich@braincorporation.com

Todd L. Hylton
hylton@braincorporation.com

August 19, 2016

## Abstract

There has been great progress in understanding of anatomical and functional microcircuitry of the primate cortex. However, the fundamental principles of cortical computation - the principles that allow the visual cortex to bind retinal spikes into representations of objects, scenes and scenarios - have so far remained elusive. In an attempt to come closer to understanding the fundamental principles of cortical computation, here we present a functional, phenomenological model of the primate visual cortex. The core part of the model describes four hierarchical cortical areas with feedforward, lateral, and recurrent connections. The three main principles implemented in the model are information compression, unsupervised learning by prediction, and use of lateral and top-down context. We show that the model reproduces key aspects of the primate ventral stream of visual processing including Simple and Complex cells in V1, increasingly complicated feature encoding, and increased separability of object representations in higher cortical areas. The model learns representations of the visual environment that allow for accurate classification and state-of-the-art visual tracking performance on novel objects.

## Introduction

Both the recurrent processing of visual information by the cortex and the temporal continuity of the visual world are an integral part of visual learning and processing in the cortex. The minimal delay of at least 100 ms between light hitting the retina and the generation of a motor output, and the general need for organisms to anticipate the dynamics of their environments indicate that *prediction* must be one of the central operating principles of sensory-motor system. At any point in time the current sensory input provides not only an update to the basis for a new prediction, but also the "correct" value against which recent prediction is compared and a network is refined. Also, since objects tend to persist in the visual field and transform coherently in time, learning by prediction generalizes their common transformations (e.g. translation, rotation, partial occlusion, change in illuminant) into higher-level representations.

Here we present *a hierarchical recurrent neural network that learns unsupervised on natural visual inputs in their natural time order, and captures increasingly high-level representations of the components of the visual scene, as information moves up the hierarchy*. Our model derives from the following principles assumed as the basis for cortical computation:

1) *Information representation and compression*: Information is compressed by learning a sparse representation [see e.g. Olshausen and Field 1996, 1997] of input to the cortical processing units (as observed in cortical area V1 macrocolumns and their associated sets of Simple cell-like receptive fields).
2) *Learning by prediction*: Learning by prediction binds temporally-adjacent representations of object features, and forms representations of visual features that are invariant to common object transformations: translation, rotation, lighting changes etc. We demonstrate that this gives rise to Complex cell-like receptive fields reminiscent of those found in V1.
3) *Context*: Lateral and top-down feedback connections provide contextual information throughout the network. This context helps to integrate spatio-temporal representations within the network and to improve predictions and learning. We demonstrate that the use of context improves the invariance of object feature representation.

Using these principles, we develop a functioning neural network model capable of classifying and tracking objects in a visual scene. The model reproduces the phenomenology of Simple and Complex receptive fields of primary visual cortex and provides predictions for higher visual areas. We have previously published a machine learning model built using these principles [Piekniewski et al. 2016]. In our current paper we show an implementation of the ideas that emphasizes correspondence with primate cortex.

The plan for the paper is as follows. In the Background Section we explain the intuition and motivations behind the model, which are derived both from features of biological cortex and from modern machine learning techniques. The details of the model implementation, the model's correspondence with cortical phenomenology, and its ability to perform tasks of tracking and object classification are given in the Methods Section. The Discussion Section presents an interpretation of our work, its relationship to the literature, and testable predictions for the function of primate visual cortex. Finally, we summarize our thoughts in the Conclusions Section, and outline directions of future research.

We hope that this model will serve as the basis both for new, more detailed models of primate cortical function and for general-purpose biologically inspired Artificial Intelligence (AI). Source code of our core model is made available at https://github.com/braincorp/ASC.

# Background

In this section we detail the inspiration for our assumed principles of cortical computation - information compression, prediction, and contextual feedback - as well as additional motivating insights from machine learning, biology, and physics.

**Information compression**: In primates, the primary visual cortex (V1) has a substantially larger number of neurons than the lateral geniculate nucleus (LGN). This is interpreted (see e.g. [Olshausen and Field 1996, 1997]) as V1 neurons forming a sparse overcomplete representation of the visual input, with receptive fields of nearby neurons being non-orthogonal and with only a relatively small fraction of neurons being active at any given instant for natural stimuli. Higher visual cortices (V2, V4) in the ventral processing stream, on the other hand, have fewer neurons than V1. This is commonly interpreted as evidence of *information compression*, where information about the visual world is represented more efficiently and irrelevant information is discarded. Observation of increasing typical receptive field sizes along the ventral pathway (V1 - V2 - V4 - PIT - AIT) has led naturally to pyramid model architectures [Fukushima 1980, LeCun et al. 1995, 1998, Riesenhuber and Poggio 1999a, Krizhevsky et al. 2012, Yamins et al. 2014] in which bottom-level units receive inputs from small patches of the visual field and top-level units receive inputs from the larger visual field as filtered by the underlying levels. Our model also has a pyramidal architecture, and we implement "information compression" via a novel sparse coding procedure.

**Prediction:** The primate visual system has access to temporal structure of the natural world, in which objects undergo stereotypic transformations such as translations, rotations, and changes in illumination. These transformations make the natural world predictable and allow the cortex to learn to represent it. Consider an edge-selective Simple cell in cortex being activated by a visual edge. Due to visual motion another nearby Simple cell, similarly oriented but with a different phase or position, will likely be active on the next time step[1]. By binding together all edge-selective Simple cells that are likely to activate on the next time step, the cortex learns to represent objects in a translationally invariant fashion. The same intuition applies not only to translation, but also to rotation, change in illumination, etc. *The model presented here learns these invariances through a prediction-driven learning technique*. When the model detects a certain sensory feature, it is likely that on the next frame a transformed feature (e.g. edge) will be also present. These original and transformed features (e.g. translation of an edge) activate different feature-encoding cells ("Simple cells" - see Methods) at their respective times. Predictive learning binds together the subsequent activities of the transforming features, thereby producing "invariant" representations ("Complex cells" - see Methods). In our networks that were trained on natural and cinematic movies, where the dominant transformation is translation, translation-invariant cells dominate the population, but we also see other types of cells exhibiting, for example, rotational invariance.

The traditional approach to creating invariant feature representations is to "engineer" them by adding together sensory representations of similar features at nearby locations [Hubel and Wiesel 1962, phase invariance as in Heeger 1992, translational invariance as in Fukushima 1980, Riesenhuber and Poggio 1999a]. For example, to have a phase-invariant orientation filter one can add squared responses of two Gabor filters 90 degrees apart in phase. In general, Complex cell responses can be engineered by wiring the Complex cell to receive input from several Simple cells that cover the desired invariance. The max pooling stage of deep neural networks is an example of this sort of engineering, where the goal is to create translation invariance in the representations of those layers by combining the response of the same filter at neighboring spatial locations via the max operator. Although effective in some applications, the engineered approach does not lend itself to generalization to the many invariances found in natural environments, which would be expected of an approach based on prediction-driven learning.

---

[1] Biologically, a time step may correspond to a cycle of gamma-rhythm, approximately 30ms.

Learning complex cell receptive fields based on spatial statistics of natural images was previously demonstrated in [Karklin and Lewicki 2009, Shan and Cottrell 2013]. Learning complex cell receptive fields based on spatio-temporal statistics of visual inputs was previously demonstrated in [Wiskott and Sejnowski 2002].

**Contextual feedback:** The primate visual system also incorporates massive cortical feedback connectivity (see e.g. [Lamme et al. 1998, Sincich and Horton 2005]). Lateral and top-down feedback connections in the visual cortex appear to communicate information that serves functions variously described as *attention* [McAdams and Maunsell 1999, Reynolds and Heeger 2009, Tsotsos 2011, Perry et al. 2015], *task definition* [Chawla et al. 1999, Li et al. 2004, Maunsell and Treue 2006], *integration* of surrounding regions [Grinvald et al. 1994, Angelucci and Bressloff 2006], and *synchronization* (see e.g. [Singer 1993, Buschman and Miller 2010]). Feedback is recognized as being primarily modulatory in nature [Lamme et al. 1998, Larkum et al. 2004]. By comparison, the vast majority of published hierarchical models of visual cortex are either feedforward [Fukushima 1980, LeCun et al. 1998, Riesenhuber and Poggio 1999a, Krizhevsky et al. 2012, Sermanet et al. 2013] or use symmetric bidirectional connections [Dayan et al. 1995, Hinton 2010]; however, see [Behnke 2003, Fukushima 2005, Grossberg 2007]. In our model abundant *top-down and lateral context is used as additional information to improve prediction*. We demonstrate that this context results in improved object classification.

**Insight from Machine Learning:** Great progress has been achieved in the last few years on AI problems such as image recognition [Krizhevsky et al. 2012, Devlin et al. 2015, Vinyals et al. 2016], language recognition [Ferrucci 2012, Mikolov et al. 2013], and games [Mnih et al. 2015, Silver et al. 2016]. This progress was, in part, driven by improvements in computational power, which provided faster training of deep networks using traditional approaches like the backpropagation of error on much larger datasets. The implementation of the backpropagation algorithm has also been improved and augmented by unsupervised pre-training or training of lower levels [LeCun et al. 2015, Schmidhuber 2015], producing faster supervised learning and better performance. In particular, deep convolutional networks have shown impressive performance on image recognition benchmarks (see e.g. [Yamins and DiCarlo 2016]). In comparison to biological cortex and to the model described in this paper, however, deep convolutional networks are typically built, optimized and trained in highly unnatural ways. First, they are trained on shuffled sets of still images rather than on temporally continuous input (video) and require supervised learning on labeled datasets. Second, they are predominantly feedforward; whereas, both the primate cortex and our model have massive lateral and top-down feedback connectivity. Third, they rely on backpropagation of errors across many levels, and consequently suffer from the vanishing gradient problem (see e.g. [Hochreiter et al. 2001, Schmidhuber 2015]). As we elaborate in detail in the Methods section below, in our model learning is *local*, which is both more biologically plausible and does not suffer from the vanishing gradient issue.

**Insight from Biology and Physics:** Objects and observer retinas move relative to each-other and change the sensory input to the visual cortex on the millisecond or subsecond time scale; yet the objects themselves persist on much longer time scales. Learning to combine sensory features that occur close together in time and space hypothetically allows to bind those features into representations of objects [Földiák 1991, DiCarlo et al. 2012]. Learning representations that are increasingly stable with respect to these identity-preserving feature changes is very likely to contribute to good object recognition both in biological systems and in AI. [Wiskott and Sejnowski 2002] have demonstrated that unsupervised learning of slow features from fast-varying sensor input data can generate Complex-like receptive fields from Simple-like ones, which allows for capturing a range of invariances from simulated visual input. There is also strong experimental evidence from primate IT recordings [Li and DiCarlo 2008, 2010] that the temporal continuity of object view serves as a teaching signal for improving the invariant representation of the object. These observations correspond closely with our expectation that *prediction* should be one of the core principles of cortical computation. Our model illustrates both the development of Complex cell-like features from exposure to natural stimuli and the ability to classify previously unseen objects using these features.

# Methods

Our model consists of a hierarchy of connected levels ("V1, V2, V3, V4" named sequentially but without implying a physiological correspondence to specific primate cortical areas), shown in Figure 1. Each level is divided into a number of tiles ("macrocolumns"), which are recurrently connected to neighboring tiles in the same level and to tiles in immediately inferior and superior levels. The number of tiles per level decreases with increasing level in the hierarchy, such that tiles in higher levels subtend increasing portions of the visual input. At the highest level ("V4" in the work presented here) a single tile subtends the entire visual space. The tiles in each level are identically constructed and consist of two layers: an input layer ("Simple cell layer") that learns to compress and represent its input as a set of activations of its cells, and an output layer ("Complex cell layer"). The activations of the Complex-layer cells predict the activations of the corresponding Simple-layer cells on the next time step. Simple cell layers receive feedforward input from Complex cell layers in adjacent tiles in immediately inferior levels. Complex cell layers receive feedforward input from their corresponding Simple cell layers, lateral feedback from Complex cell layers in neighboring tiles at the same level, and top-down feedback from the Complex cell layers in adjacent tiles in immediately superior levels. The lateral and feedback connectivity provides context, i.e., extra information that serves to improve the prediction of the future Simple cell activations, but the feedback itself is not predicted by the Complex layer. The prediction of future activation requires learning temporally persistent "causes" of Simple cell activations, which correspond to persistent features of the visual field such as objects. This highly-interconnected, recurrent network is intended to create a consistent, efficient representation of the spatio-temporal structure of the input.

## Architecture

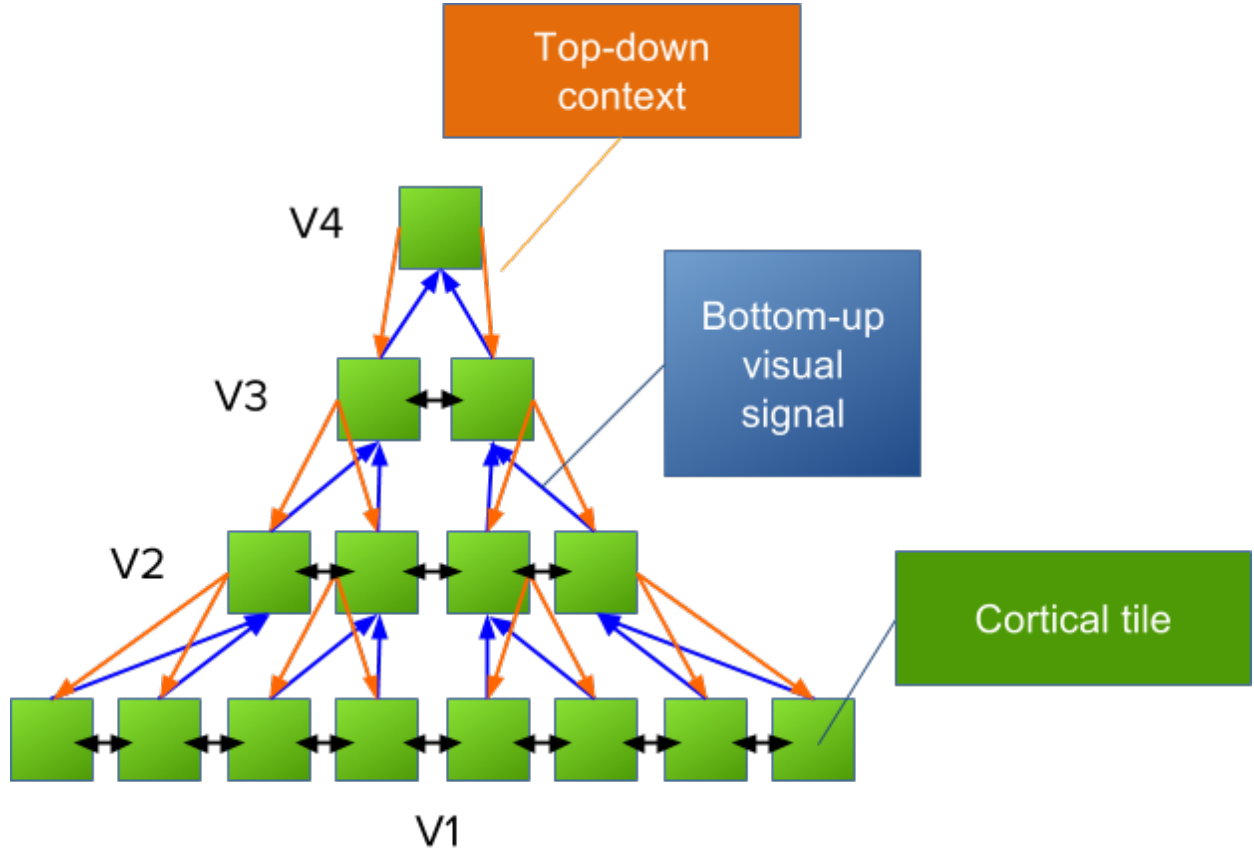

**Figure 1**: Network architecture. Each hierarchical level is made up of identical tiles. A tile in each level receives feedforward input from several tiles of lower level (blue arrows), feedback input from higher-level tile (orange arrows), and lateral input from neighboring tiles in the same level (black arrows).

The model processes videos of 80x80 pixels; this resolution was chosen because it results in a model 4 levels deep with 64 V1 tiles of size of 10x10 pixels (larger models are possible with a corresponding increase in computational resources). The model is composed of 4 levels, arranged in a hierarchy, with labels V1-V4. Each level spans the whole visual field without overlap or gaps between the tiles. Level V1 is composed of 64 (8x8) tiles, with each tile receiving 100 (10x10) pixels as input. Level V2 has 16 (4x4) tiles and each tile receives input from 4 (2x2) V1 tiles. V3 has 4 (2x2) tiles and each tile receives input from 4 (2x2) V2 tiles. Level V4 has one tile, which receives input from 4 (2x2) V3 tiles. Hence, each tile (in V2 and above) receives input from 4 tiles beneath it and the total number of tiles decreases by a factor of 4 for every level in the hierarchy. The top-down feedback connections project down to the lower tiles from which the higher tile receives inputs. Tiles within the same layer receive lateral connections from their immediate neighbors (4 neighbors for tiles in the center of the layer, 3 for those on the edge, and 2 for those on the corners). Each tile has a Simple cell layer and a Complex cell layer. We used 400 cells per Simple layer and 400 cells per Complex layer in each tile, except where noted otherwise. The Simple cell layer uses a form of Sparse Coding described below to learn a dictionary of features that represent the feedforward input. The

Complex cell layer receives the output of the Simple cell layer plus all lateral and feedback connections (from the previous time step) in order to predict the next Simple cell layer output. The output of the Complex cell layer is the prediction of the Simple cell layer's next output. This prediction is the output of the tile, which is fed to the higher, lower and lateral tiles for feedforward, feedback and lateral connectivity. In order to accelerate learning as well as to make the models more computationally efficient, connection strengths (weights) were shared between tiles within the same level.

The number of simulated neurons (typically 68,000) is, of course, much smaller than the number of the neurons in biological visual cortex. The model therefore has both lower visual acuity (results presented are for 80x80 pixel visual input) and lower overcompleteness of the cortical feature representations[2].

The model was trained by viewing 3 back-to-back nature documentaries (approximately 300K frames, 2.8 hours of video) repeated 10 times for a total of 3M frames. All layers and levels were trained simultaneously. The learning is unsupervised - each tile within the network learns locally to predict its next input.

The model is implemented in Python, with heavy optimization using Cython in order to allow multithreading, and runs at approximately 20 frames per second during training and approximately 50 frames per second after training using 20 threads on a 40 hyperthread core 2.9GHz Intel Xeon workstation.

## Simple cell layer learning rule

To learn sparse features we use a variation on Sparse Coding [Olshausen and Field 1997] that we call Adaptive Sparse Coding (ASC) described below. Sparse coding uses a dictionary of feature vectors to reconstruct the input. Each cell in the Simple cell layer learns a set of input weights (connection strengths) that result in its activation when its preferred feature is present in the input. The goal of Sparse Coding is to find linear activations of these feature vectors which result in good reconstruction of the input (i.e., low reconstruction error) while also keeping the number of non-zero activations low. The traditional approach to sparse coding is to use a sparsity parameter, λ, which balances reconstruction error and the level of sparsity (normally the $\ell_1$ norm of the activations). By contrast, ASC minimizes reconstruction error while keeping the sparsity level fixed (i.e. the $\ell_0$ norm, the number of non-zero values in the response vector often referred to as "the number of active cells", is fixed). We note that while finding the global minimum in reconstruction error while keeping the $\ell_0$ norm constant is computationally very expensive, an approximate minimum can be found in reasonable time and in fact is often faster than traditional Sparse Coding[3]. Also, as compared to traditional Sparse Coding, ASC offers the advantage of automatic normalization. Traditional Sparse Coding will either generate no response or a very non-sparse response to complex patterns very far from its learned features (e.g., white noise). Additionally, either input vectors are normalized to have unit variance (which destroys input scale information) or as the contrast of the image is varied the number of active cells in the traditional Sparse Coding can change drastically. Because ASC constrains the number of active cells, it does not suffer from too few or too many responses for any given input, with the tradeoff that reconstruction error can vary substantially depending on the complexity of the input.

---

[2] A single neuron in the model may be treated as a "representative" neuron of a cortical minicolumn or of a group of neurons with similar receptive fields.

[3] Unlike the SC algorithm, which loops until convergence, the ASC algorithm loops only until the $\ell_0$ norm constraint is satisfied.

Specifically, ASC is trying to optimize the following:

$$\mathbf{a} = \mathrm{argmin}_a(||\mathbf{y} - \mathbf{Da}||_2^2 + \lambda ||\mathbf{a}||_1), \text{ with } \forall\ \mathbf{a}_i >= 0 \text{ while finding } \lambda \text{ that results in } ||\mathbf{a}||_0 = N$$

Here, **D** is the dictionary of receptive fields represented by the connection weights of the Simple cells within a tile, **y** is a single input frame to the tile, and **a** is the sparse activations vector of those Simple cells. The first term is the reconstruction error, i.e. the square of the $\ell_2$ norm. The second term is the $\ell_1$ penalty on **a** that imposes sparseness.

The optimization algorithm is shown in Algorithm 1. In general, increasing K, the number of cells, or N, the number of active cells, improves reconstruction error. However, for a fixed K, increasing N beyond roughly K/4 results in V1 receptive fields lacking the typical Gabor-like structure, poor object classification and poor tracking (not shown). Increasing K improves the representation but at a significant computational cost. The computational complexity is approximately $O(NK^2)$. The scaling rule for $\lambda$ (line 13) is a form of binary search; though the exact scaling values used are arbitrary, they work well. The average $\lambda$ scale, s, is an optimization to accelerate finding the appropriate value of $\lambda$, since starting from a known good average point relative to the maximum activation of **z** (lines 3 and 5) often allows the sparsifying loop to exit sooner; however, once learning is disabled, the average $\lambda$ scale, s, is made constant. Unless stated otherwise, K=400, N=70, and T=25, where T is the maximum number of iterations per time step. The learning rule was inspired by the excellent review on sparse coding by [Mairal et al. 2014]. Note, while Algorithm 1 is constructed for images, other modalities or inputs can easily be used by modifying line 2.

**Algorithm 1:** Adaptive Sparse Coding.
**Require:** RGB image **x** in $\mathbb{Z}^m$, dictionary **D** in $\mathbb{R}^{m \times K}$, maximum number of iterations T, target number of active cells N, average $\lambda$ scale $s_0$.

1:   **a** ← 0;
2:   **x** ← **x** - 127.5;  // make 0 mean gray
3:   **z** ← **xD**;  // co-activation vector
4:   **E** ← $\mathbf{D}^T\mathbf{D}$;  // autocorrelation of the dictionary
5:   $\lambda$ ← max(**z**) * $s_0$;
6:   t ← 1;
7:   **while** $||\mathbf{a}||_0$ != N **and** t <= T
8:     **for** i = 1,…,K **do**
9:       d ← $\lfloor \mathbf{a}[i] + \mathbf{z}[i] - \lambda \rfloor$ - **a**[i];  // where $\lfloor\ \rfloor$ is half-rectification, d is the delta for a[i]
10:     **if** d != 0
11:       **z**[i] ← **z**[i] - d**E**[:,i];
12:       **a**[i] ← **a**[i] + d;
13:   $\lambda$ ← $\lambda$((1.0+2.0/t) **if** $||\mathbf{a}||_0$ > N **else** (1.0-0.75/t));
14:   t ← t + 1;
15:   s ← $0.999 s_0$ + $0.001 \lambda$ / max(**z**);
16:   **return a,** s;

During learning, as a performance optimization, the weights in the dictionary **D** are not updated on every time step; instead the changes are accumulated and the weights are updated at intervals that increase with training. The starting interval is 1000 frames and each subsequent interval is increased by a factor of 1.1. Thus the updates occur with intervals of 1000, 1100, 1210, etc. which correspond to steps 1000, 2100, 3310, etc. Learning of dictionary weights was accomplished using a slightly modified block coordinate descent [Mairal et al. 2014] with momentum, Algorithm 2.

**Algorithm 2:** Dictionary update rule using block coordinate descent.
    **Require:** initial dictionary $\mathbf{D}_0$ in $\mathbb{R}^{m \times K}$, input images **Y** in $\mathbb{R}^{m \times n}$ for current update interval, sparse activations **A** in $\mathbb{R}^{K \times n}$ for current update interval, initial correlation between the inputs and activations $\mathbf{B}_0$ in $\mathbb{R}^{m \times K}$, initial autocorrelations $\mathbf{E}_0$ in $\mathbb{R}^{K \times K}$.
1: $\mathbf{D} \leftarrow \mathbf{D}_0$;
2: $\mathbf{B} \leftarrow (\mathbf{B}_0 + \mathbf{Y}\mathbf{A}^T)/2$;
3: $\mathbf{E} \leftarrow (\mathbf{E}_0 + \mathbf{A}\mathbf{A}^T)/2$;
4: **for** $i$ = 1,...,K **do**
5:   $\mathbf{D}[:,i] \leftarrow (\mathbf{B}[:,i] - \mathbf{A}[:,i]\mathbf{D}^T) / (\mathbf{E}[i,i] + 0.0000001)$;
6:   $\mathbf{D}[:,i] \leftarrow \mathbf{D}[:,i] / (||\mathbf{D}[:,i]||_2 + 0.0000001)$;
7: **return D**, **B**, **E**.

Figure 2 shows an example of what learned features look like in V1. A dictionary of size K=400 is visualized as a 20x20 grid of 10x10 image patches; each image patch is a single feature (Simple cell receptive field).

In order to ensure that certain cells don't dominate the output response, all sparse activations are normalized to have on average cell $\ell_2$ norm over time before being passed to the Complex layer, as shown in Algorithm 3. This normalization can be thought of as a form of homeostasis or whitening. To simplify the downstream algorithms a constant-activation cell was appended to all sparse, Simple cell, responses and thus for the Complex layer algorithms J=K+1 is used instead of K.

**Algorithm 3:** Sparse, Simple cell, activation normalization.
    **Require:** unnormalized sparse activation $\mathbf{a}_0$ in $\mathbb{R}^K$, autocorrelations **E** in $\mathbb{R}^{K \times K}$ from output of dictionary update.
1: $\mathbf{v} \leftarrow \text{diag}(\mathbf{E})$;
2: **for** $i$ = 1,...,K **do**
3:   $\mathbf{a}[i] \leftarrow \mathbf{a}_0[i] / (\mathbf{v}[i]^{0.5} + 0.0000001)$;
4: **return** [**a**, 1]; // normalized sparse, Simple cell, response.

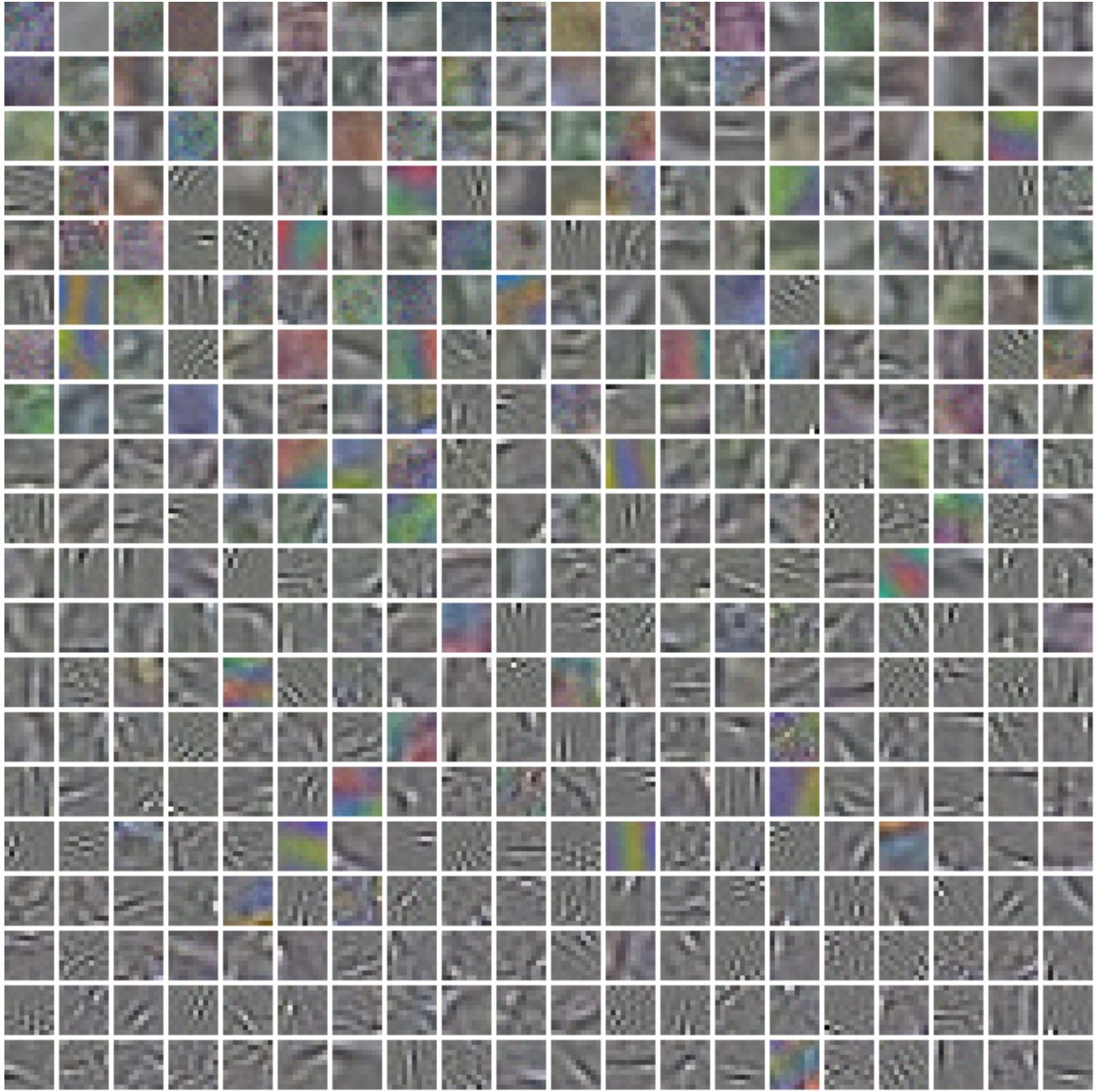

**Figure 2**: V1 Simple Features visualized as 10x10 pixel patches on a grid 20x20 (dictionary size, K=400). Each patch shows the pattern of input to which the particular cell is sensitive, often referred to as the Receptive Field or weights of the cell. The entire collection of patches is the Dictionary of a tile containing the cells. Color and orientation features of varying spatial frequency and spatial extent are learned by watching nature documentaries.

## Complex cell layer learning rule

The objective for learning in the Complex layer is to use the current (sparse) response of the Simple layer and context from lateral and feedback connections to predict the next response of the Simple layer. As a consequence of this construct, each Complex cell has an associated single Simple cell for which it is predicting; specifically, Complex cell #1 predicts the activity of Simple cell #1, Complex cell #2 predicts the activity of Simple cell #2, etc. To do this we employed a single layer Perceptron with a half-rectified linear activation function, Algorithm 4. In order to learn the Complex layer weights (**C** matrix) with the half-rectified linear activation function, we employ stochastic gradient descent using a slow learning rate and a weak $\ell_2$ penalty term for most weights and a strong $\ell_2$ penalty term for "self" connections, Algorithm 5. Because the current response of any cell *i* of the Simple layer is in general strongly correlated with its next response (i.e., the likelihood of a cell having similar activation in adjacent video frames is very high), the weights from cell *i* Simple to cell *i* Complex must be treated differently: these special weights are the "self" connections, which lie on the diagonal of the **C** matrix. We found that without a strong penalty for self connections, the weight matrix would be nearly an identity matrix which did not provide very good Complex receptive fields or good classification performance. A penalty factor of approximately 0.5 or larger works well but 0.9 was chosen because it caused the self weights to be approximately the same magnitude as the non-self weights. Unlike Simple cell learning, Complex cell learning requires continuous weight updates so as to prevent uncontrolled positive feedback between updates. A slow learning rate is also required to keep the feedback dynamics stable: faster learning rates result in oscillations and transient but detrimental positive feedback. In general, oscillations[4] and pathological weight growth are not maintained during learning because they are not useful in predicting the next input: weights responsible for these pathologies are therefore diminished by the learning rule.

**Algorithm 4:** Complex cell activation.
  **Require:** Complex cells weight matrix **C** in $\mathbb{R}^{7J \times J}$, previous normalized Complex response $c_0$ in $\mathbb{R}^J$, current normalized Simple sparse response **a** in $\mathbb{R}^J$, normalized lateral responses $l_0$ in $\mathbb{R}^{4J}$, normalized feedback response $f_0$ in $\mathbb{R}^J$, learning step *t*.
1: $p_0 \leftarrow [a, c_0, l_0, f_0]$; // full context used for Complex layer prediction, in $\mathbb{R}^{7J \times 1}$
2: $c \leftarrow \lfloor p_0 C \rfloor$; // where $\lfloor \ \rfloor$ is half-rectification
3: **return c**;

---

[4] We applied eigenvalue analysis to the low order recurrent connections and determined that they are strictly stable (the real part of all eigenvalues are less than 1).

**Algorithm 5:** Complex cell learning rule.
    **Require:** initial Complex cells weight matrix $C_0$ in $\mathbb{R}^{7J \times J}$, previous normalized Complex cell responses $c_0$ in $\mathbb{R}^J$, current normalized Simple sparse response $a$ in $\mathbb{R}^J$, future normalized Simple sparse response $a_1$ in $\mathbb{R}^J$, normalized lateral response $l_0$ in $\mathbb{R}^{4J}$, normalized feedback response $f_0$ in $\mathbb{R}^J$, learning step $t$.

1: $C \leftarrow C_0$;
2: $r \leftarrow 1 / (10000 + t / 10.0)$;  // slow down learning over time
3: $C \leftarrow (1.0 - 0.00001r) \cdot C$;  // weak $\ell_2$ penalty
4: $\text{diag}(C) \leftarrow (1.0 - 0.9r) \cdot \text{diag}(C)$;  // strong "self" $\ell_2$ penalty
5: $p_0 \leftarrow [a, c_0, l_0, f_0]$;  // full context used for Complex layer prediction, in $\mathbb{R}^{7J \times 1}$
6: $s \leftarrow [1 \text{ if } c_0[i] > 1 \text{ else } 0 \text{ for } i = 1,...,J]$;  // a boolean vector where $c_0$ is positive
6: $d \leftarrow (a_1 - c_0) \cdot (s + 0.01)$;  // in $\mathbb{R}^J$, 0.01 is to ensure the cells don't stop responding
7: $T \leftarrow p_0 d$;  // in $\mathbb{R}^{7J \times J}$
7: $T \leftarrow T / \max(1.0, \max(||T||_1) + 0.0000001)$;  // clip the gradient elements to 1
5: $C \leftarrow C + rT$;
6: **return** $C$.

In order to ensure that certain cells don't dominate the output response, all Complex activations are normalized to have on average cell $\ell_2$ norm over time, see Algorithm 6. This normalization can be thought of as a form of homeostasis or whitening.

**Algorithm 6:** Complex cell response normalization
    **Require:** unnormalized Complex cell responses $c_0$ in $\mathbb{R}^J$, initial average variance $v_0$ in $\mathbb{R}^J$, learning step $t$.

1: $r \leftarrow 1 / (10000 + t / 10.0)$;
2: $v \leftarrow (1 - r) \cdot v_0 + c_0^2$;
3: **for** $i = 1,...,J$ **do**
4:     $c[i] \leftarrow c_0[i] / (v[i]^{0.5} + 0.0000001)$;
5: **return** $c$;  // normalized Complex response.

The diversity of learned Complex features are shown in figures 3, 4, 5, 6, and 7. Figure 3 shows Complex cells that have learned to respond to multiple phases and positions of the same orientation. Learning phase and positional invariance is the most common feature of the Complex layer. Figure 4 demonstrates that the Complex layer also learns to combine Simple cells with similar color selectivity to have broader hue selectivity than seen in the Simple layer. Figure 5 shows the Complex layer learning to combine multiple phases for the same orientation. Figure 6 demonstrates that there are some Complex cells that might be sensitive to multiple orientations suggestive of rotational or curvature invariance. Finally, Figure 7 shows an unexpected population of Complex layer cells that are selective to a combination of color and orientation. Since colors have much lower spatial frequencies than luminance the learned color features are predominately full field with a few color edges. Thus these cells may be a type of "color-stop" which encodes the sharp edge of a color region Similar structure is observed in V2, shown in Figure 11.

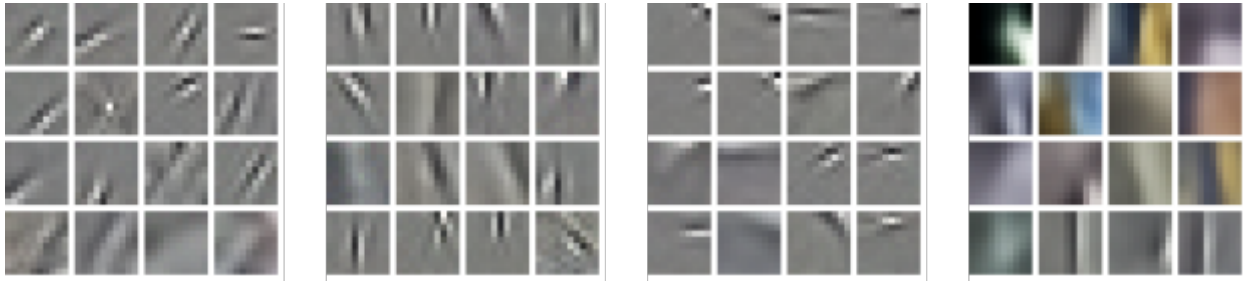

**Figure 3:** Orientation selectivity. Four V1 Complex receptive fields that are orientation selective. The 16 V1 Simple receptive fields that contribute most to a single Complex cell are shown on a 4x4 grid, for each Complex cell. Shown from left to right are Complex receptive fields with a purely luminance (all gray) preference for diagonal, vertical, horizontal, and a low spatial frequency vertical with some color selectivity.

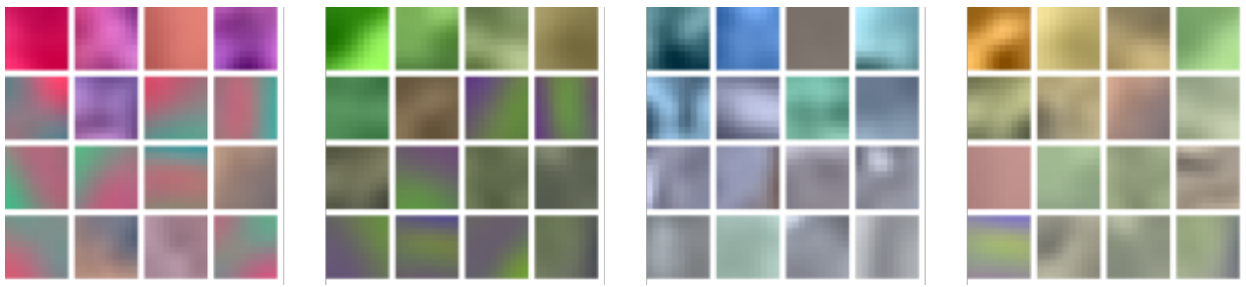

**Figure 4:** Color selectivity. Four V1 Complex receptive fields that are hue selective. The 16 V1 Simple receptive fields that contribute most to a single Complex cell are shown on a 4x4 grid, for each Complex cell. Shown from left to right are "red-ish", "green-ish", "blue-ish" and "yellow-ish" Complex cells.

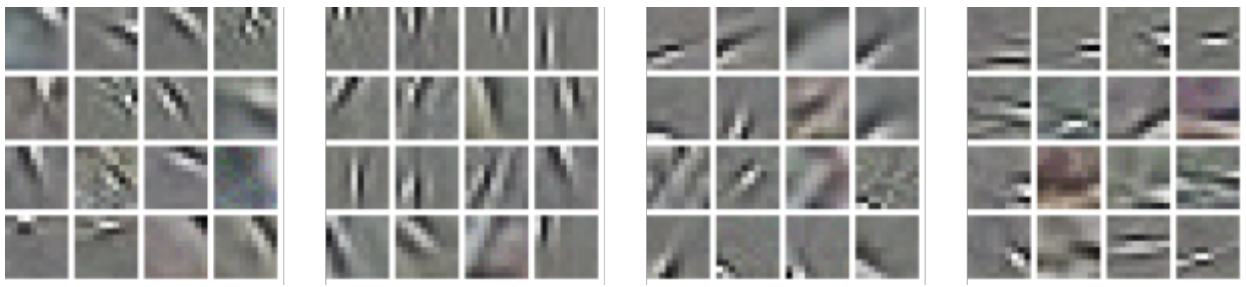

**Figure 5:** Broad spatial frequency tuning. Four V1 Complex receptive fields that demonstrate broad spatial frequency tuning for the same orientation. The 16 V1 Simple receptive fields that contribute most to a single Complex cell are shown on a 4x4 grid, for each Complex cell.

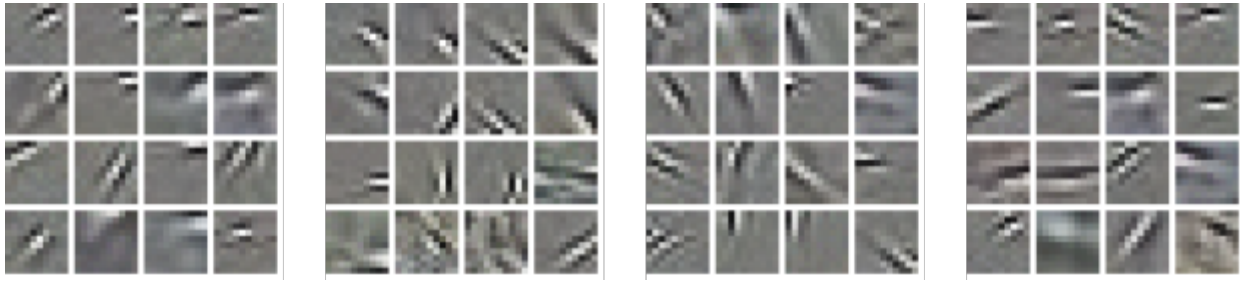

**Figure 6:** Range of orientations prefered. Four V1 Complex receptive fields that demonstrate selectivity to a range of orientations. The 16 V1 Simple receptive fields that contribute most to a single Complex cell are shown on a 4x4 grid, for each Complex cell. **Left**: orientation preference spans approximately 45 degrees, **middle and right:** cells demonstrate selectivity for all orientations.

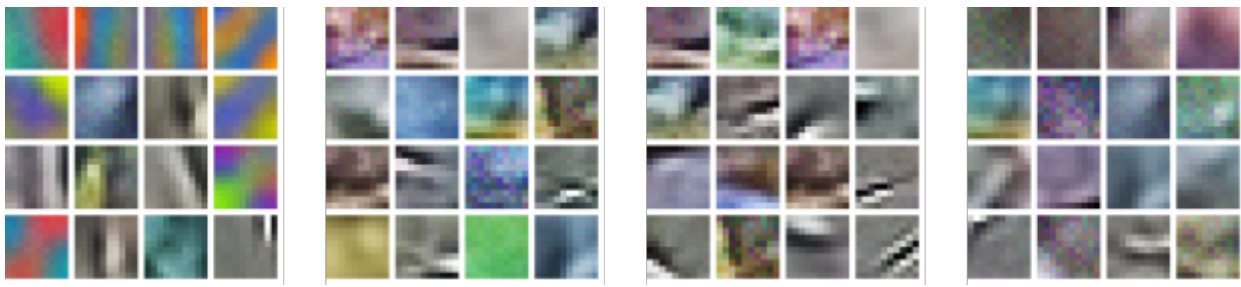

**Figure 7:** Unexpected combined color and orientation preference. Four V1 Complex receptive fields that unexpectedly demonstrate selectivity to a combination of color and orientation. The 16 V1 Simple receptive fields that contribute most to a single Complex cell are shown on a 4x4 grid, for each Complex cell. **Left**: cell that prefers vertical luminance and and nearly unoriented color edges, **middle and right:** cells prefer oriented luminance and nearly full field color.

One may argue that Figures 6 and 7 could be a result of undertraining, however, the cells shown in Figure 7 in particular converge to these states early in training and do not deviate much with additional training. Further, these cells are a minority of the population. It should also be noted that for primates the majority of the cells in V1 have unknown responses [Olshausen and Field 2004].

## White Noise Analysis

In order to evaluate V1 Complex receptive fields in a way that is commonly done in biology we performed Spike-Triggered Covariance analysis [Rust et al. 2005] using 500,000 random 10x10 RGB frames; note that we did not remove the Spike-Triggered Average before performing the covariance analysis, see Figure 8. These results demonstrate that the cells in the Complex layer truly are Complex by more traditional measures, in that they have multiple significant excitatory features with different phases and spatial frequencies but otherwise similar preference.

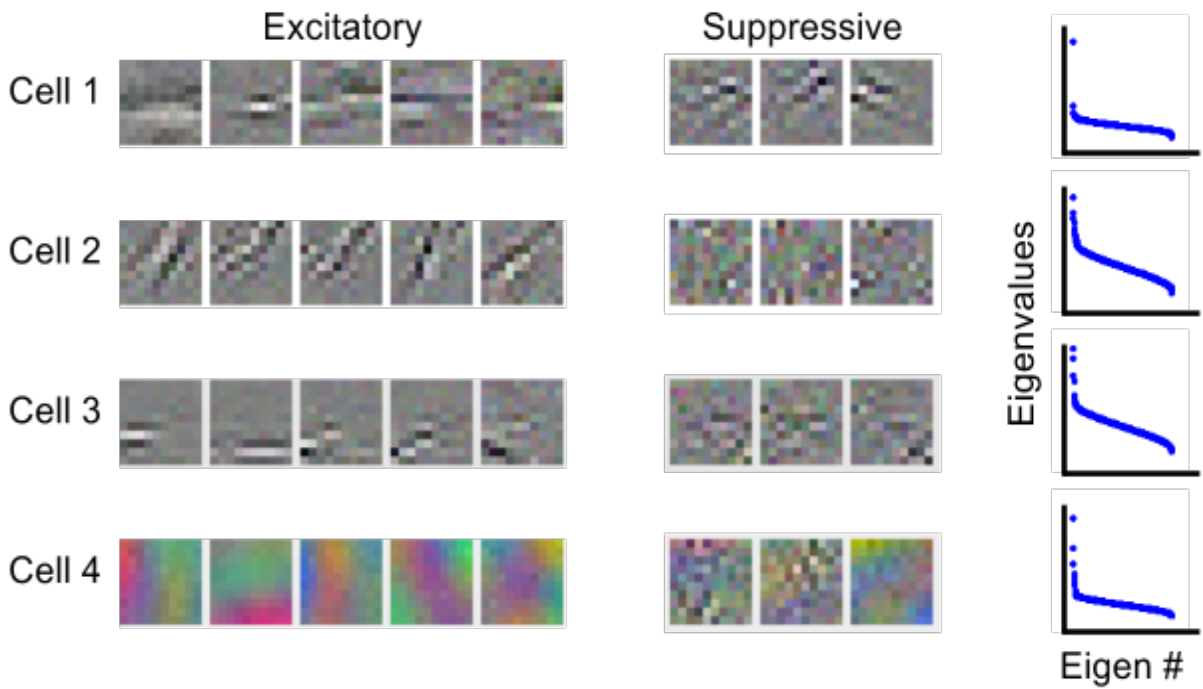

**Figure 8**: White Noise Analysis. Four example cells from layer V1 Complex are shown. **Left**: the top 5 excitatory components. **Middle**: 3 most suppressive components. **Right**: the eigenvalue plots. These results demonstrate that all of these cells have multiple excitatory and suppressive feature dimensions consistent with them being considered Complex. cell 1 prefers horizontally oriented luminance features; the first excitatory component is much stronger than others (see eigenvalue plot). Cell 2 prefers diagonal orientations; several strong excitatory components are present. Cell 3 prefers horizontal but doesn't have a dominant feature like cell 1. Cell 4 is a Complex color cell that has several strong excitatory features. For all cells, the suppressive features differ from the excitatory ones in their location and/or preference.

## V2 features

Each V2 tile receives feedforward input from 4 (2x2) V1 tiles. As a result, for a model with 400 Simple cells per tile, every V2 Simple cell receives 1600 feedforward inputs. To visualize a selected V2 Simple cell receptive field, we analyze each of the 4 (2x2) V1 input tiles independently. In each V1 tile we determine the 9 (3x3) V1 Complex cells that contribute most to the response (9 highest weights) of the selected V2 Simple cell, then render the V1 Simple cell that corresponds to those 9 V1 Complex cells. Figure 9 demonstrates V2 orientation selective cells that have relatively similar preferences across all 4 (2x2) input tiles. Figure 10 demonstrates V2 color selective cells that have relatively similar preferences across all 4 (2x2) input tiles. Figure 11 demonstrates more complicated V2 receptive field preferences.

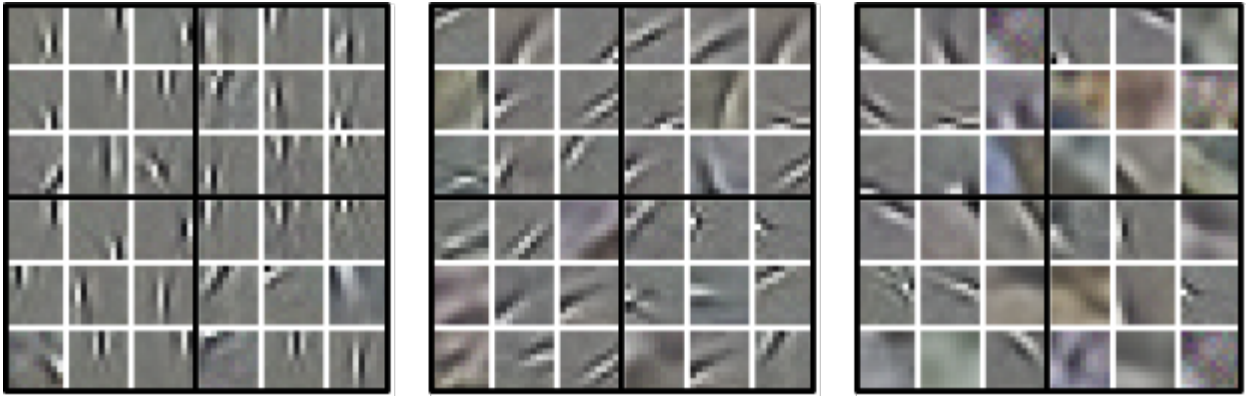

**Figure 9**: Similar orientation tuning across space. Three V2 Simple cells are shown. For each V2 Simple cell, four V1 tiles that feed into it are shown on a 2x2 grid (black outlined boxes). Within each box, the 9 V1 Complex cells within the tile that contributed most to the V2 cell's response are visualized as the associated V1 Simple cell RF that the Complex cell predicts. **Left:** All 4 tiles predominantly prefer vertical orientation. **Middle:** All 4 tiles predominantly prefer forward diagonal orientation. **Right:** All 4 tiles predominantly prefer backward diagonal orientation though with some small amount of color selectivity.

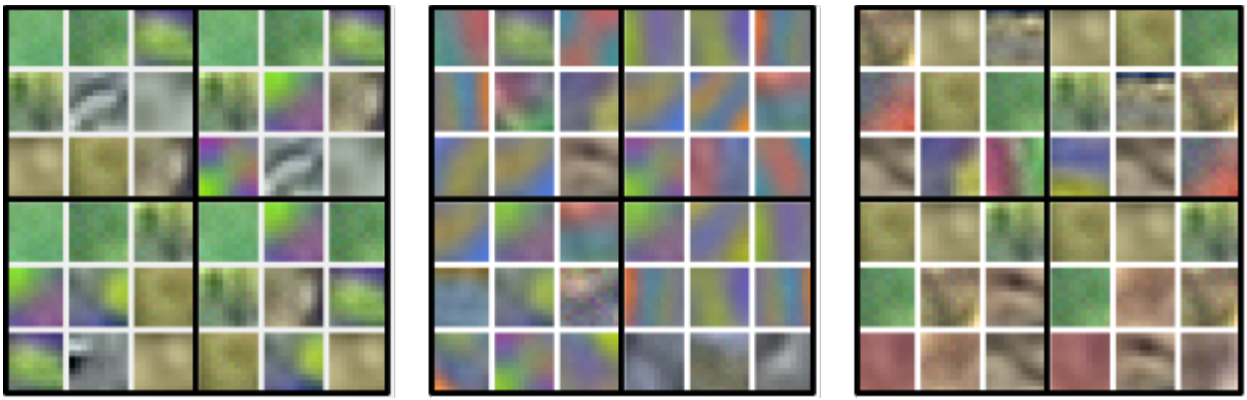

**Figure 10**: Similar color preference across space. Three V2 Simple cells are shown. For each V2 Simple cell, four V1 tiles that feed into it are shown on a 2x2 grid (black outlined boxes). Within each box, the 9 V1 Complex cells within the tile that contributed most to the V2 cell's response are visualized as the associated V1 Simple cell RF that the Complex cell predicts. **Left:** All 4 tiles predominantly prefer green. **Middle:** All 4 tiles prefer mixtures of red-green and blue-yellow edges. **Right:** All 4 tiles predominantly prefer yellow.

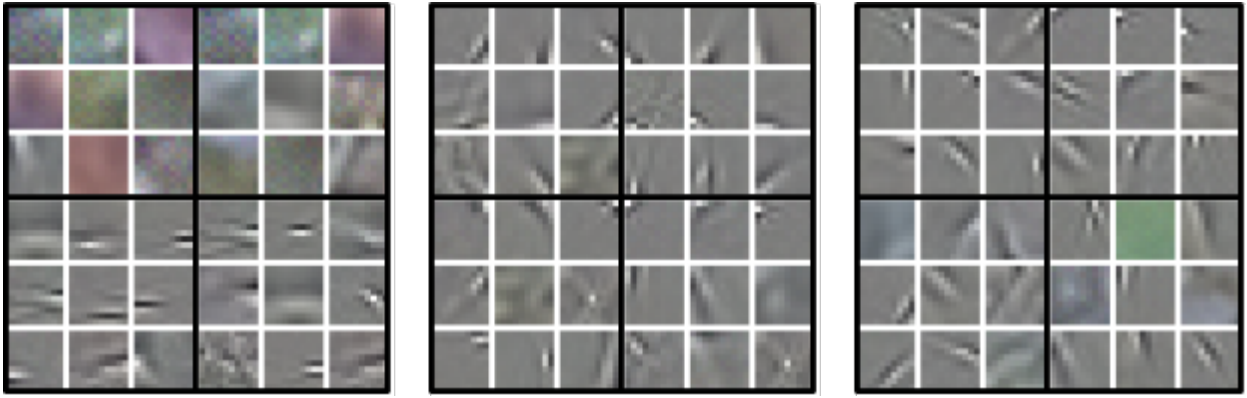

**Figure 11**: Complicated V2 preferences. Three V2 Simple cells are shown. For each V2 Simple cell, four V1 tiles that feed into it are shown on a 2x2 grid (black outlined boxes). Within each box, the 9 V1 Complex cells within the tile that contributed most to the V2 cell's response are visualized as the associated V1 Simple cell RF that the Complex cell predicts. **Left:** top 2 tiles predominantly prefer full field color and bottom tiles predominantly prefer horizontal orientation. **Middle and Right:** Similar to the V1 cells shown in Figure 6, each tile seems to respond to a wide range of orientations, possibly representing curvature or rotation.

## High order features

For cells in high visual cortices, objective quantification of the receptive fields poses substantial challenges. Here we used two approaches: (a) finding the natural stimuli that produce the largest activations of a cell in top level (V4) of the model, and (b) performing artificial stimulus optimization to maximize the activation of this cell. To find the natural stimuli that produce the largest activations of a V4 cell, a selectivity measure was first defined: $s_i = a_i/\text{sum}(a)$, where $a_i$ is the activation of cell $i$ for any particular input frame, and the sum is over all cells for the current frame. The training natural videos were used as input. To reduce similarity between highest-activation frames, we only recorded an s value for every 100th frame. For a given cell i, the 9 frames with the highest s are shown in figure 12 (left). Further, an optimization procedure was developed whereby an image basis set was formed by taking the top 1000 eigenvectors of the training video frames and using this 1000 dimensional space to optimize (maximize) the selectivity measure for a single random cell using Nelder-Mead optimization, figure 12 (right).

The receptive field structure revealed by this analysis demonstrate that cells in Level 4 of the model have converged to predominantly complicated features that represent common full-frame statistics.

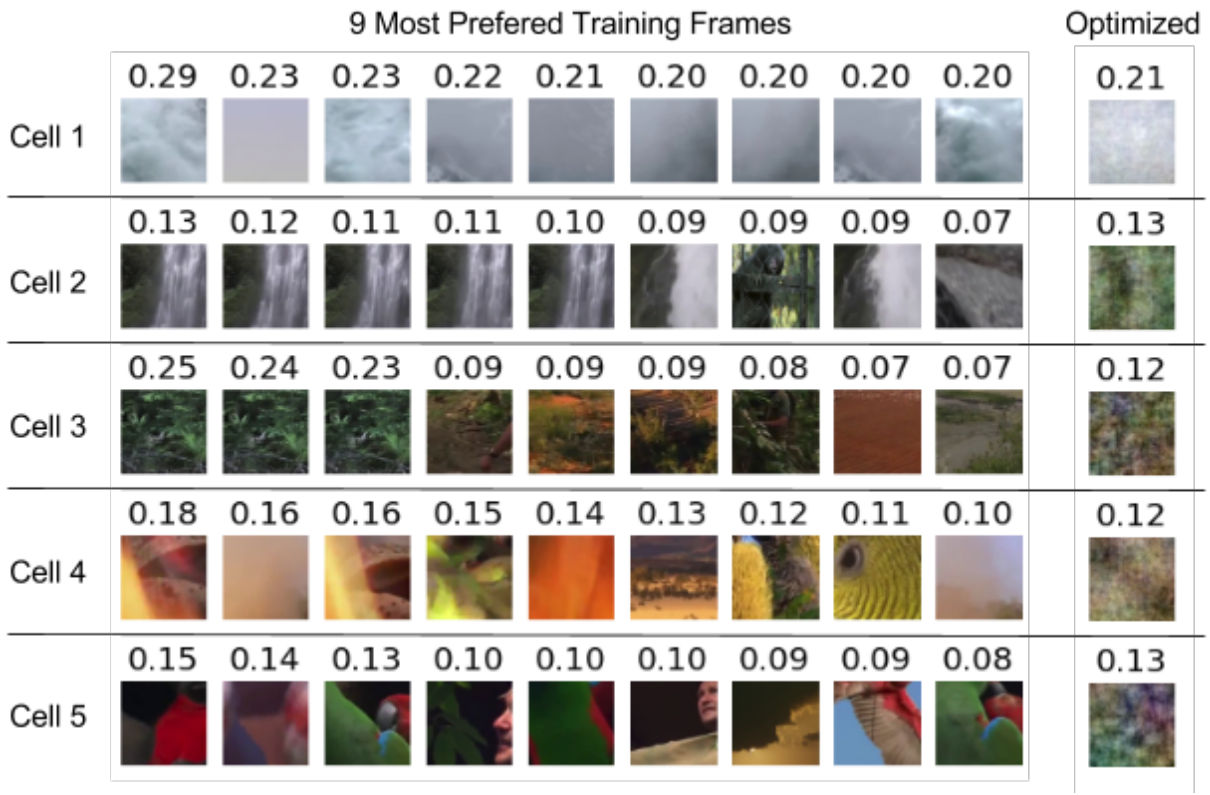

**Figure 12**: From the V4 Simple layer, 5 cells were randomly selected. For each cell the 9 frames that the cell was most selective for are shown on the left. Shown on the right is the image that maximized the selectivity when performing Nelder-Mead optimization on a 1000 dimensional basis set (see text). Shown above each image is the cell's selectivity, $s_i$, value for that image. Cell 1 clearly prefers white, though there is some texture selectivity since there were several cells in the model that preferred white but with different textures (not shown). Cell 2 prefers vertical green structures. Cell 3 prefers predominantly high spatial frequencies along the red-green color axis. Cell 4 seems to prefer yellow diagonal structure. Cell 5 in general prefers low spatial frequency red vs green.

## Motion selective features

We have also tested the ability of the model to learn motion selective features. Specifically we let the input vector **x** be the concatenation of 2 to 5 sequential frames instead of the single frame implementations reported above. This additional temporal information in the input enables the network to learn motion selective features in the cells within its Simple and Complex layers. Under all tested configurations the network successfully learned highly direction selective cells in the V1 Simple layer; approximately 25% of the population became direction selective with different direction preferences, spatial frequencies, speeds and receptive field sizes (Figure 13). An additional 10% of the population wasn't direction selective but flips preference over time; the remaining 65% have the same preference over time. For a closer look at the diversity of temporal dynamics of the V1 Simple cell preferences see Figure 14. Direction selective cells were also found in the Complex layer; but unlike the Simple layer cells, these direction selective Complex layer cells showed evidence of additional invariances related to phase, multiple spatial frequencies and speed (not shown). This result suggests a potential for pattern-motion selectivity [Adelson et al. 1983]. We note that others have also demonstrated learning motion using sparse coding [Cadieu and Olshausen 2008].

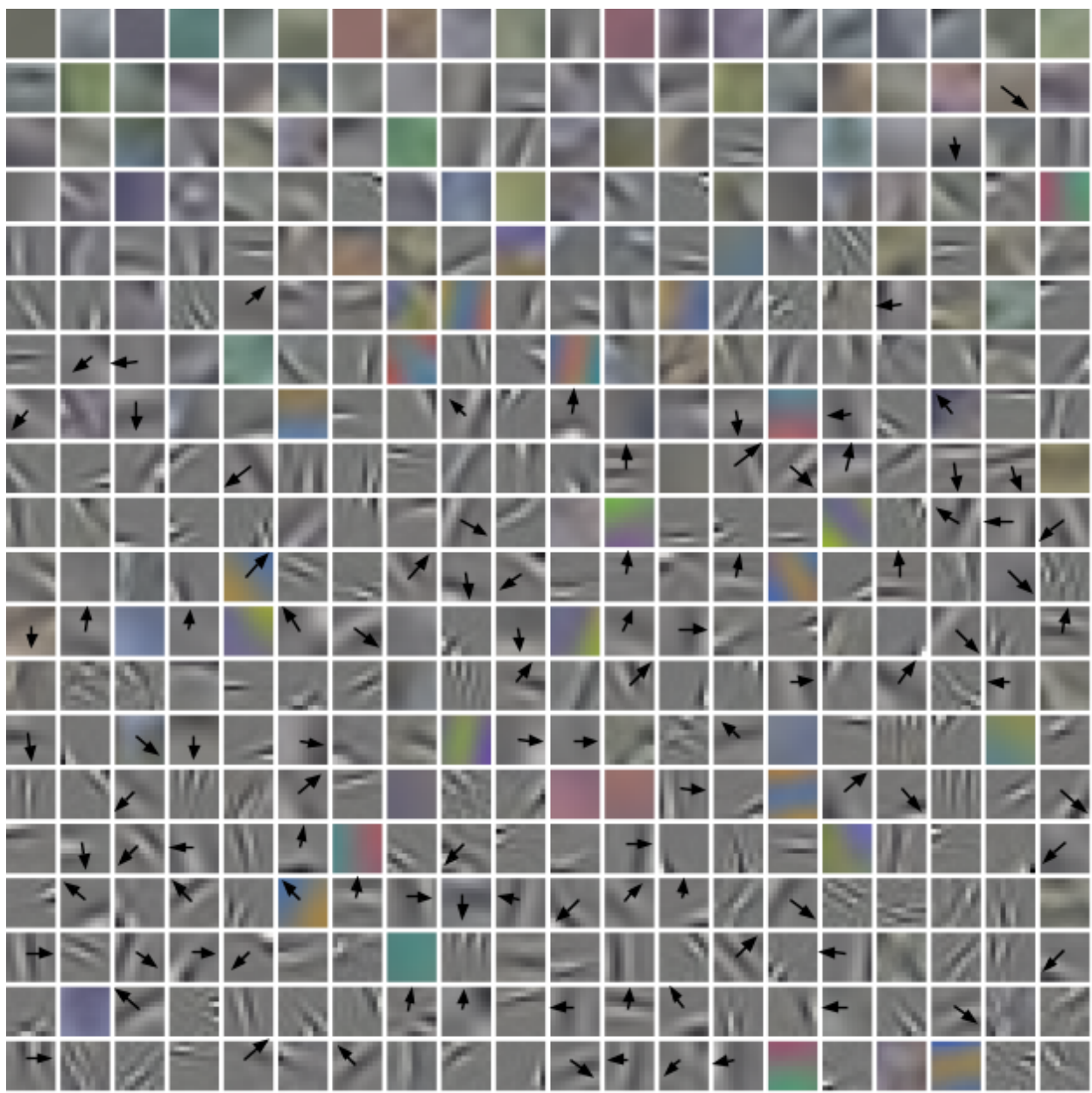

**Figure 13**: Direction selective cells. 400 V1 Simple features visualized as 10x10 pixel patches on a grid 20x20. The model learned on video from nature documentaries using 3 video frames concatenated together as input. For visualization purposes only, the first frame (of 3) of the motion receptive field are shown. There are approximately 100 cells that are direction selective, their direction of preference is indicated by an arrow. Additional temporal dynamics beyond motion were observed, see Figure 14.

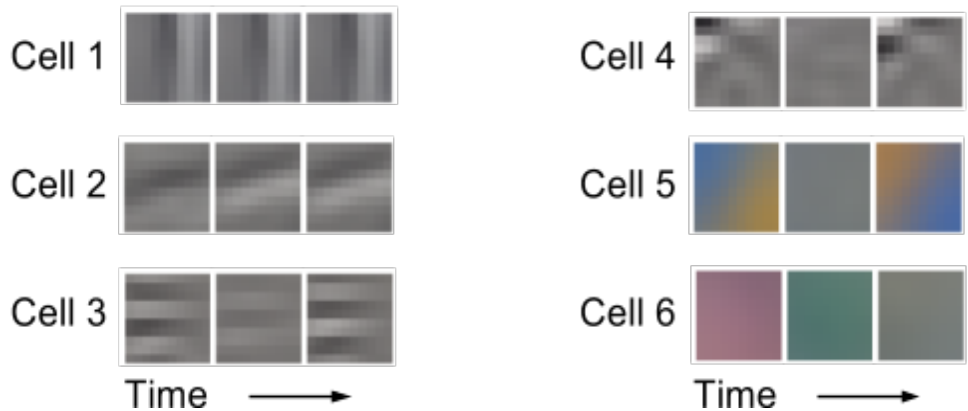

**Figure 14**: Six V1 Simple cells are shown with a diversity of temporal dynamics. Cell 1 does not change preference over time and represents the majority of cells. Cell 2 prefers up and to the left motion. Cell 3 prefers upward motion. Cell 4 flips polarity over the 3 frames. Cell 5 flips color polarity with a slight motion preference for up and to the left. Cell 6 flips color preference from pink to green-blue followed by gray.

## Image Classification

To quantify the benefits of predictive learning and of the contextual feedback in the visual hierarchy, a classification task was used. The choice of the object classification task was motivated by the need to quantify how invariant the representations are to translation and scaling. We therefore constructed a stimulus that uses objects that are randomly scaled and translated to determine how well a classifier would generalize to a new set of scales and positions. If the network indeed learns to represent useful invariant features, then the representations of visual objects would be easily separable by a shallow classifier that uses network activity as an input.

We reiterate that the hierarchy itself learns unsupervised. It binds together the features of objects it sees, but it does not know the object labels. A separate shallow neural network is trained to associate the hierarchy activations with the object labels. This classifier is trained and tested after the hierarchy itself has finished training on natural movies, and all of the connection weights in the hierarchy have been fixed. The objects the classifier is trained and tested on are never used in unsupervised training of the hierarchy.

In detail, the procedure was as follows. First, the model was trained on natural videos as described above, and upon completion of training all learning was disabled. Next a single layer perceptron classifier with hyperbolic tangent activation function was trained on the outputs (V1 Simple, V1 Complex, V2 Simple, etc. up to V4 Complex) of the model. The images to be classified were generated from 41 static objects, Figure 15, that were randomly positioned and scaled with black backgrounds. Classification was done per tile, which corresponds to a 10x10 pixel region of visual space for V1 or 20x20 pixels for V2, etc. Due to this small aperture and low image resolution, this classification task is extremely difficult and thus classification performance is not expected to be near 100%, however, relative improvement in classification is informative.

The supervised classifiers used to evaluate the classification performance are single layer perceptrons trained independently on all layers, but shared between tiles of the same layer. Since all tiles in the same layer have the same features, a classifier trained for one can be used on all. Furthermore, the outputs of all tiles can be used as training examples for the same classifier.

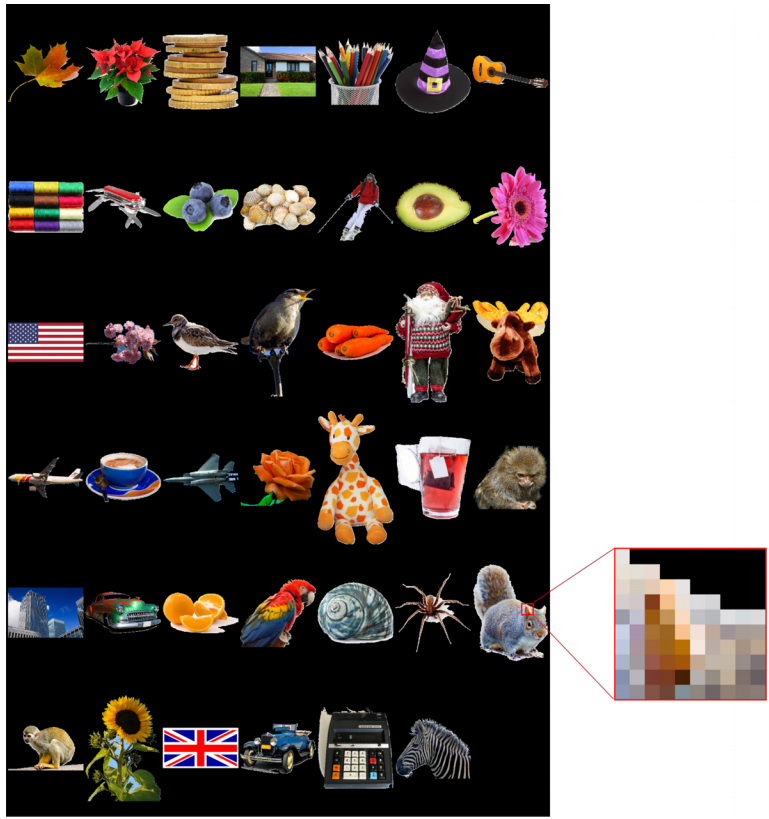

**Figure 15**: **left**: Images of 41 objects used for classification performance testing. **right**: example of 10x10 subimage used for classification by a single V1 tile.

Different dictionary sizes, and network architectures were evaluated. Figure 16 shows the performance of 3 different sized networks (200, 400, 900) on this classification task. The data suggest that there is a substantial improvement in classification between K=200 and 400 cells and a small improvement in classification between K=400 and 900 cells.

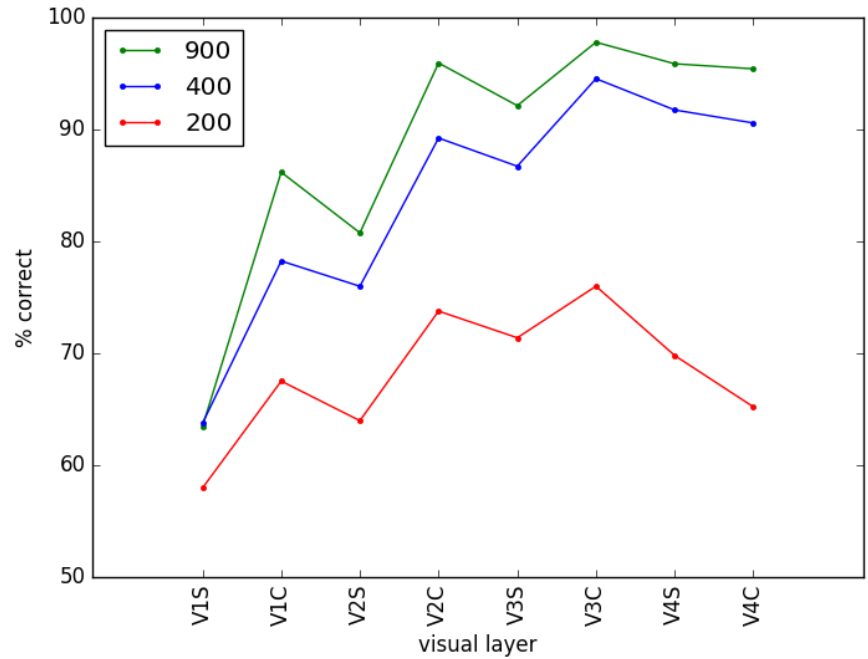

**Figure 16**: The classification performance of 3 different models that differ in dictionary sizes: K=200 (red), 400 (blue), and 900 (green) cells. The classification performance of single layer perceptrons trained independently on activations of every layer in these models were evaluated on classifying 41 objects that were randomly scaled and positioned within the visual field. Chance level of classification is 2.44%.

Figure 17 shows the performance of the model on the object classification task. Performance of the full model is shown in blue. In red we show the performance of a reduced model in which the Complex cell layer did not receive lateral and top-down contextual feedback, both during training and during the tests. In green we show the performance of a further reduced model which only had Simple cell layers in all four levels. Clearly, the model that was trained and is operating with contextual feedback has formed object representations that are substantially more linearly separable. The best performance is achieved in V3 Complex layer of the network with feedback. This performance is better than that of any layer of the network without feedback. We also note that feedback selectively improves the performance of the Complex layers (the layer that receives the feedback). V1 Simple has no access to feedback and thus the classification performance based on V1 Simple activity is identical with and without feedback. Somewhat counterintuitively, we find that (at least in some networks trained) the performance of the level 4 is lower than the performance of level 3 Complex cell layer (see blue line in Figure 17). We note, however, that level 4 is the top level and is a single tile, and thus receives neither top-down nor lateral feedback. Further, due to the lateral connectivity, level V3 tiles actually have access to the entire visual field, whereas level V4 receives input from the entire visual field but only after compression. This compression may explain the drop in classification performance between level V3 and level V4.

To determine the benefit of having a Complex layer, we also compared these models to a model lacking a Complex layer (Figure 17 green curve), specifically the Simple cell layer becomes the output of the tile; this model is reminiscent of the stacked autoencoder network of [Hinton and Salakhutdinov 2006]. This model underperforms for all layers after V1 compared to our models with a Complex layer. Additional evidence for the benefit of the Complex layer comes from looking at the relative gain between Simple and Complex layers within the same level. For both the red and blue curves in Figure 17 there is a general improvement from the Simple to Complex layers (except level V1 for red and level V4 for both red and blue).

The difference between the classification performance of the Complex layer and the higher level Simple layer shows the effects of Compression (e.g. the decrease observed in the blue curve in Figure 17 between V1C and V2S, V2C and V3S, etc.) and it also shows the benefit of having a larger visual field (e.g. the increase observed in the red curve between V1C and V2S, V2C and V3S, etc.).

These image classification results provide strong evidence that our model learns invariances as well as useful visual features for object classification.

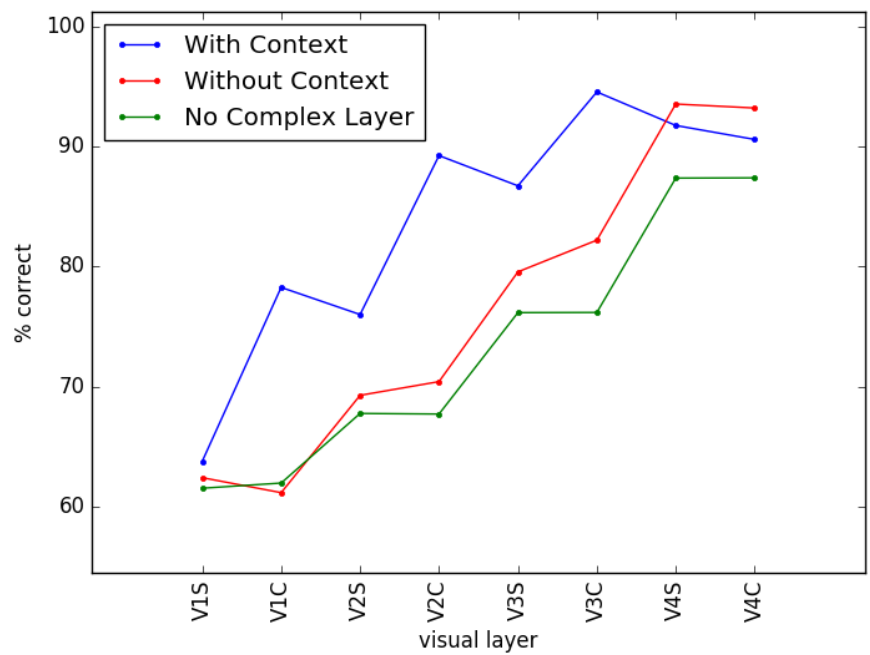

**Figure 17**: Comparison with reduced models. Results for three models are shown: a model with context/feedback (blue, duplicate of the blue curve in Figure 16); a model without context/feedback (red); and a model without Complex layers (green). All three models have dictionary size K=400 cells. The classification performance of single layer perceptrons trained independently on all layers in these models were evaluated on classifying 41 objects that were randomly scaled and positioned within the visual field. First, note that there is a significant benefit of using feedback in classification since for layers after V1S the blue curve is substantially higher than red. Second, the highest classification performance occurs in Layer V3 Complex. For the model without a Complex layer (green curve), the Complex layers (V1C, V2C, etc.) are repeaters of the corresponding Simple layers. Chance level of classification is 2.44%.

## Tracker implementation

To test the generality of our vision model we tested it while performing a tracking task. In order to adapt the vision model for tracking we added object-location heatmap classifiers to each level of the vision model. In each video clip on which the tracking performance was evaluated, the first frame was used for priming the tracker, that is, for training the classifiers to generate a heatmap estimating the location of the target object. Namely, the first frame and a bounding box representing the object to track were scaled and translated in many configurations to train the heatmap classifier. Classifiers were trained *de novo* on the priming frame of every tracking video clip, as target objects differed between videos. Once trained, the classifier weights were fixed. Each classifier was then used to generate a heatmap based on the activity of all cells in the corresponding level in the hierarchy. For example, V1 had 64 (8x8) tiles with 400 Complex cells per tile; these 25.6K cells were used as input to a classifier that then generated an 80x80 output image where large positive pixel values indicate presence of the target and values near 0 indicate its absence. The V2 heatmap classifier used 16(4x4) tiles with 400 cells per tile for a total of 6.4K cells, and so on. Overall, the tracking algorithm combined the 4 classifier heatmaps together in order to estimate the true target location and to form a bounding box. Since the input resolution of the model is relatively low (80x80) compared to the input video, the tracker runs in a "windowed" mode (active tracking); specifically, the vision model views a cropped subregion of the full image, and the subregion is centered on the estimated target location of the input image. The subregion is dynamically shifted during the course of tracking, according to the tracker's estimation of the target position.

Example heatmaps for each level are shown in Figure 18. As can be seen in Figure 18, the cells in the top level (V4) of the model are at least as informative with regard to the target object position as the cells in the bottom level (V1). Superficially this seems counterintuitive, as model V4 has only one tile whereas V1 has 8x8 = 64 tiles. Furthermore, in the oversimplified traditional view the cells at the top level of the ventral stream encode the object identities but not the object positions. However, both in primates [Hong et al. 2016] and in the current model, the cells in the top level of the ventral stream do encode enough positional information to localize the object of interest within the visual field. Please also see [Riesenhuber and Poggio 1999b].

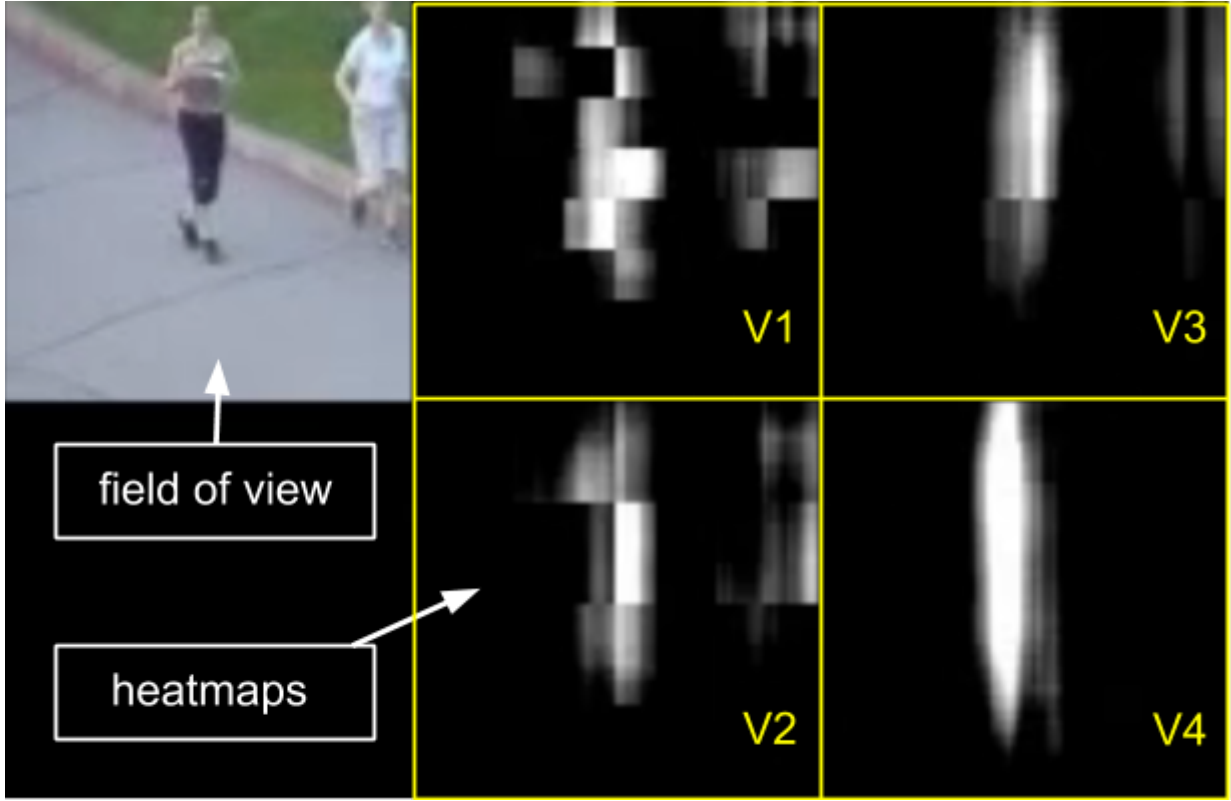

**Figure 18**: Target object location heatmaps (middle and right panels) computed from activity of each of the four levels of the hierarchy V1, V2, V3, and V4 for the current frame (top left panel). In this video the tracker was primed on the jogger who is currently on the left. Heatmaps estimate the location of the target from the activity of the tiles in the associated level as a function of location in the input frame; white indicates higher likelihood. Interestingly, V4 does not respond to the distractor (the jogger in white) at all and V3 responds to the distractor only minimally which is consistent with the image classification results where levels V3 and V4 have the best classification performance.

Tracking results are shown in figures 19 and 20 where our model is compared to a state-of-the-art tracker (STRUCK, [Hare et al. 2011]). The datasets used for testing were the Tracker Benchmark from [Wu et al. 2013] (Figure 19) and our own Green Basketball dataset [Piekniewski et al. 2016] (Figure 20). Our complementary paper on the Predictive Vision Model [Piekniewski et al. 2016] contains a wealth of additional information on tracking and comparisons to other trackers. Briefly, Success measure $S(\theta)$ is defined as the fraction of frames in which the overlap (ratio of area of the intersection to the area of the union) of the ground truth bounding box and the tracker bounding box is greater than a given argument $\theta$ (from 0 to 1). Better trackers have a larger area under $S(\theta)$ curve; a perfect tracker would have the area under $S(\theta)$ curve equal to 1. Accuracy is a resolution-independent measure that quantifies tracking with regards to both true positives (target is present and detected) and true negatives (target is absent and not detected). Namely, it measures the number of frames in which the center of tracking box lies inside the ground truth bounding box plus the number of frames in which the target is absent and the tracker also reports that is is absent, normalized by the total number of frames in the video excluding the priming frame. For figures 19 and 20, the Accuracy measure (left panel) was varied as a function of the bounding box size (scale), such that the bounding box was artificially scaled by the horizontal factor to visualize how close the tracker's bounding box is to the ground truth, in units of the size of the ground truth bounding box (1 means unscaled).

The tracking results demonstrate that our visual hierarchy is capable of tracking on par with state-of-the-art trackers. Note, the Tracking Benchmark dataset has been published for a while and the STRUCK tracker has been highly tuned for this data. Conversely, STRUCK was not tuned for our Green Basketball dataset, but because our model was in part developed while testing performance against it, our model may reflect optimizations for it. We note, however, that our model is intended primarily to model cortex, not to serve as a tracker.

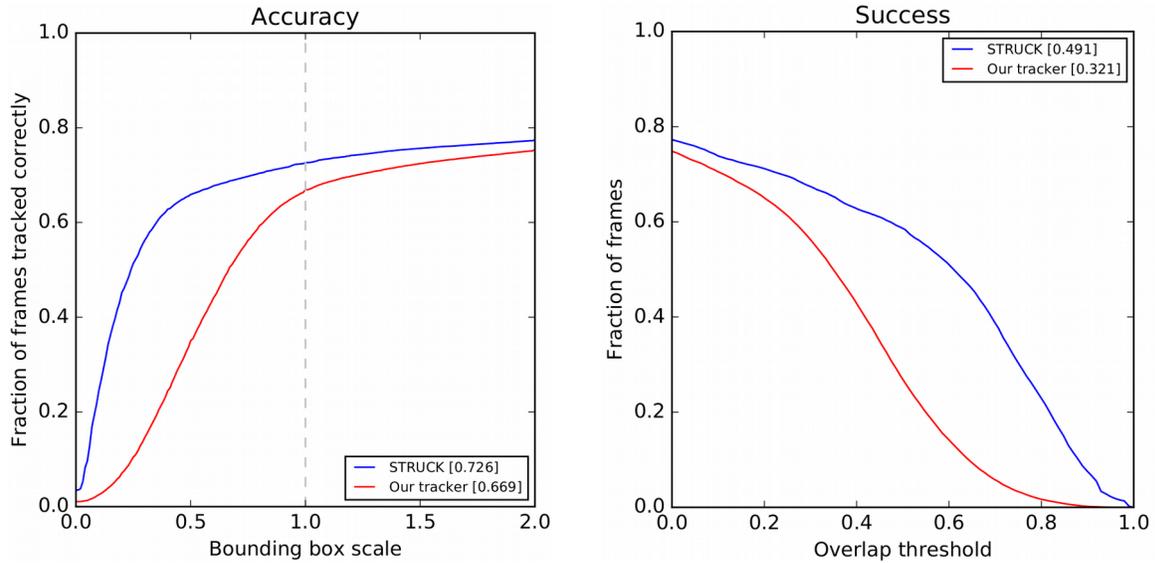

**Figure 19**: Tracking performance on Tracking Benchmark dataset [Wu et al. 2013]. Accuracy (left) and Success (right) plots are shown. The better the Accuracy, the more the curves are to the upper left. The better the Success, the more the curves are to the upper right. Our tracker (red) is close to but slightly underperforming compared to another state-of-the-art tracker such as STRUCK (blue).

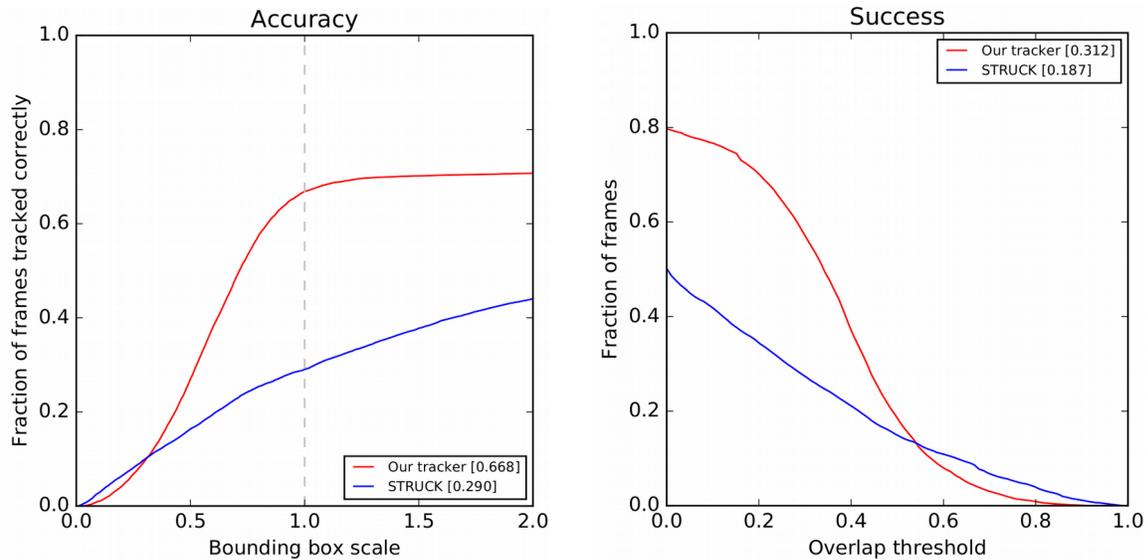

**Figure 20**: Tracking performance on our Green Basketball dataset [Piekniewski et al. 2016]. Accuracy (left) and Success (right) plots are shown. The better the Accuracy, the more the curves are to the upper left. The better the Success, the more the curves are to the upper right. Our tracker (red) substantially out performs STRUCK (blue).

# Discussion

This Section has two main parts. In the first part, we discuss the cortical prediction mechanisms and models, experimental evidence of cortical prediction, and possible experimental approaches to verifying or falsifying the hypotheses implemented by the current model (as compared to models in literature). In the second part of this Section we discuss the alternative implementations of our model.

## Temporal prediction

So far, the use of prediction in the models of the sensory cortex has been focused on information compression and prediction error. Proposals of cortical predictive coding can be found in the literature (see e.g. [Carpenter and Grossberg 1987, Mumford 1992, Rao and Ballard 1999, Bastos 2012, Clark 2013]). Indeed, the predictable part of the input signal carries no additional information, and therefore may be "predicted out". These works suggest that a higher cortical area predicts the activity of a lower one. This allows the lower area to receive the prediction as top-down feedback and to transmit forward the prediction error. A detailed treatment of the proposal that a higher cortical area predicts the activity of a lower one in a Bayesian Framework is found in [Bastos et al. 2012]. Further, instead of being "predicted out", the predictable part of the signal can simply be removed (e.g. by a spatial or temporal derivative) see [Druckmann et al. 2012, Clark 2013]. Parasol retinal ganglion cells, for example, encode spatial and temporal differences in luminance, which makes sense [Srinivasan et al. 1982] since luminance tends to be highly correlated in both space and time.

By contrast, our model uses prediction as a means of learning temporal persistence. In reality, both the predicted representation of the input (as in our model) and the prediction error have utility for the higher areas. In our model prediction is carried out locally: the Complex cell layer in the cortical area predicts the same-area Simple cell layer activity, and this prediction results in learning invariant representations. The prediction error, on the other hand, is useful (a) as a highly compressed proxy for the input information, and (b) as a "surprise" signal that may be used for driving bottom-up attention when and where the prediction error is large. In primates it seems physiologically plausible that both the predicted signal and the prediction error are being conveyed and utilized: the prediction predominantly via direct feedforward cortico-cortical connections, and the prediction error predominantly via pulvinar. Experimental evidence (see e.g. [Desimone and Duncan 1995]) suggests the role of pulvinar in attention consistent with the above hypothesis.

## Invariant responses vs. temporal stability

Invariant responses are more temporally stable than non-invariant ones, but temporal stability of responses does not imply invariance. Other groups have generated temporally stable sparse coding algorithms such as LCA [Rozell et al. 2007] that force stable responses in time in an attempt to learn better features. Prediction provides a mechanism for learning temporally stable representations, unlike LCA which enforces the feature stability ad hoc. Furthermore, prediction automatically learns appropriate time scales for integration of features (different features may have different time courses), whereas LCA enforces a single time scale.

## Experimental evidence

Several lines of experimental evidence are either consistent with or give indirect support for our model's architecture and for its basic hypotheses on the roles of prediction and feedback.

1. **The model architecture is consistent with visual cortex anatomy.** Feedforward input predominantly targets layer 4 and can activate or silence layer 4 spiny stellate cells. Long-range feedback (both lateral and from higher visual areas) predominantly targets apical tuft dendrites of layer 2/3 and 5 pyramidal cells and provides a mostly modulatory effect, consistent with the contextual role of feedback.
2. **The model is consistent with V1 electrophysiology.** While there is a continuum of V1 cell responses from Simple to Complex [Priebe et al. 2004, Yeh et al. 2009, Fournier et.al. 2011], the relative fraction of Complex cells is lower in the input layer (layer 4C) and higher in the output layer (layer 2/3) of primate V1 [Ringach et al. 2002]. Further, [Yeh et al. 2009] observed that V1 layer 4 cells do not change their receptive fields nearly as much as layer 2/3 cells when the input stimulus statistics is changed. This is, again, consistent with the present hypothesis that contextual feedback influences the prediction (i.e., the responses of the layer 2/3) of V1 but not the responses of the Simple cell layer of V1.
3. **The model is consistent with speed-accuracy tradeoff.** "Single-pass processing" observed in experiments that test for the shortest latency of response [Thorpe et al. 1996, Hung et al. 2005] suggests that visual processing can be carried out based on feedforward input alone; however, allowing more time for recurrent processing increases the performance on the task (speed-accuracy tradeoff [Kirchner and Thorpe 2006]). This is consistent with our model: prediction can be carried out without context or with irrelevant context, and as recurrent processing brings the context in agreement with the input the prediction accuracy increases. Backward temporal masking experiments ("interruption masking", see e.g. review [Enns and Di Lollo 2000]) lead largely to the same conclusion [Lamme et al. 2002, Bacon-Macé et al. 2005].
  4. **The model is consistent with inactivation experiments.** Inactivation of areas V2 and MT has been observed to *reduce* the response of V1 neurons to visual stimulation of their RF center (see e.g. [Angelucci and Bresloff 2006] and the references therein). This seems contrary to the hypothesis [see Mumford 1992, Rao and Ballard 1999, Bastos et al. 2012] that V2 predicts V1 and V1 feeds forward the prediction error. Indeed, when V2 is inactivated and makes no prediction, the prediction error should be large and V1 readout layers should be *more* active. Our hypothesis, on the other hand, is consistent with the observations because it does not predict dramatic changes in V1 activity when V2 is inactivated (i.e. when top-down feedback is removed); but does predict small changes in receptive fields of V1 layer 2/3 neurons.

## Testable predictions of our model

An interesting testable prediction may allow to disambiguate between the two predictive-processing hypotheses: prediction-error and local-prediction (our model). Under [Mumford 1992, Rao and Ballard 1999, Bastos et al. 2012] hypothesis (prediction-error), V2 predicts V1 activity and V1 layer 2/3 feeds forward the error of that prediction. In our model, on the other hand, V1 predicts its own activity (Complex cell layer predicts the activity of the Simple cell layer), and Complex cell layer (biologically - layer 2/3) feeds forward the prediction rather than the prediction error. Under our hypothesis it would seem likely that adding pixel noise to a highly predictable natural visual stimulus should not strongly change the firing rate of V1 layer 2/3; while under the hypothesis that V1 layer 2/3 reports the prediction error the firing rate of V1 layer 2/3 pyramidal cells should increase as the noise power increases. Also, according to our hypothesis, activity of individual V1 layer 2/3 pyramidal cells should predominantly correlate with the predictable component of the input signal; whereas under the hypothesis that V1 layer 2/3 reports the prediction error, the activity of individual V1 layer 2/3 pyramidal cells should predominantly correlate with the noise component of the input signal.

Of course, both our model (where inter-area feedforward connections carry the prediction) and the prediction-error models (where inter-area feedforward connections carry the prediction error) are oversimplifications of the actual primate cortical function, and as such may generate unrealistic predictions (e.g. in one extreme that overall activity level of V1 layer 2/3 does not depend on predictability of the LGN input to V1, or in another extreme that V1 layer 2/3 falls silent when LGN input to V1 is highly predictable). A realistic model of the cortex would be found somewhere between these two extremes, with the cortical area output having both prediction and prediction error channels. A key distinction is, however, whether the higher area predicts the activity of the lower area, or whether output layer of each area predicts the activity of the input layer of the same area. These two hypotheses are not necessarily mutually exclusive: it is possible that prediction (and predictive learning) occurs in both sets of connections.

## Alternative implementations

The architecture described in the Methods section was in large part motivated by its correspondence with selected features of primate cortex (see above). However, there are potentially many other architectures that would also satisfy our core hypotheses of compression, prediction and context. For example, the Predictive Vision Model (PVM) described in our complementary paper [Piekniewski et al. 2016] is an example of an alternative architecture using the same hypotheses. The PVM model combines prediction and compression into a single step using a multilayer perceptron with sigmoid activation functions as a cortical tile. The input to the perceptron is the feedforward connections from the lower-area cortical tiles plus lateral and feedback connections used as context. The output is the predicted next-step feedforward input as represented by a middle layer of the perceptron (configured to have fewer units than inputs), thus providing a compressed representation used by higher tiles as input and by lower and lateral tiles as context. PVM does not use weight sharing but otherwise has a crystalline hierarchical structure similar to the one described in this paper. Many other implementations are likely possible, but we believe that they all share the following.

- Although the objective of prediction may differ in different implementations (e.g. predicting the next input frame, several frames ahead, input from another sensory modality, or an externally applied training signal), prediction at every level is required to provide robust training signals throughout the network and, thereby, to mitigate problems like vanishing gradients.
- Feedforward and feedback (including lateral) connections must play different roles: the feedback signal provides context for improved prediction of the feedforward input, but the feedback signal itself is not predicted. This segregation provides for stability of learning - if prediction of feedback were allowed then the system would likely be unstable.
- A directed, acyclic graph is required for the feedforward information processing. In other words a strictly feedforward path for the information treated as input must be included.
- While a strict pyramidal hierarchy is not required, the number of tiles must decrease along the input / feedforward path so that input information is compressed as it moves through the network.
- At least one non-linearity per level is required. For the model in this paper there are two nonlinear stages: sparse coding and half-rectification.

In addition to alternative architectures, predictive networks might be trained in many ways that we have not presented here. An interesting alternative to using the next time step input as the teaching signal is to instead use an independent signal, possibly even a different modality, for teaching. In our related PVM model [Piekniewski et al. 2016], the Readout Signal used to generate the heatmap for tracking is an example of what an alternative teaching signal may be. The PVM model predicts both the Primary Signal (input, $P_{t+1}$) and the Readout Signal ($M_{t+1}$). Generalizing beyond that which is presented in the PVM model, this alternative teaching signal, instead of being an additional signal that must be predicted, could serve as the only teaching signal for a tile ("unit" in PVM nomenclature). We assert that all of the beneficial properties of the recurrent hierarchical network model still apply under this alternative teaching signal model.

# Conclusions

We have developed and implemented a prediction-based hierarchical vision model capable of unsupervised learning of the spatial and temporal regularities of the real world. The model implements learning of sparse overcomplete feature representations, predictive learning of invariances from the continuity of natural video, and improved prediction by use of substantial recurrent feedback (lateral and top-down).

The model reproduces key features of the primate visual system: emergent Simple and Complex cells in the primary visual cortex, representation of increasingly complicated features in higher levels of the hierarchy, and improved separability of object representations in higher levels. The model generates testable predictions, e.g. for the dependence of the activity of individual cells and populations of cells with the level of noise in the visual input.

We have furthermore demonstrated that the model successfully performs tasks such as object classification and tracking of moving objects. Tracking performance is on par with the state-of-the-art commercial tracking software, even though our model is not designed as a dedicated tracker but rather as a general-purpose, neuromorphic, recurrent, learning vision model with contextual feedback. The model allows tracking to be one of the tasks learned. The core learning of the model is unsupervised, followed by brief supervised training on a specific task.

The model can easily be expanded to increase its biological realism and thus to elucidate or predict specific aspects of function of primate sensory cortex. Additionally, the model can be scaled up for use in a general-purpose AI.

# Future Directions

**Multimodal integration**: The principle of context-based prediction directly generalizes to cross-modal prediction or prediction of multimodal stimuli. The sound of a bird or the smell of a flower are predictive of their visual appearance, and vice-versa. Visual, vestibular, tactile, etc. stimuli predict each-other and are integrated to provide a consistent internal representation of the body state and position in the environment. Thus, context provided by one modality may improve prediction in the sensory hierarchy of another modality. Furthermore, joint prediction (multimodal integration) may be more accurate than separate predictions within each modality.

**Objectness**: The above is true for the "what" and "where" streams of the visual processing providing context to each-other. For example, the thick stripes of V2 represent information regarding motion and stereopsis (i.e. "where"), whereas thin and pale stripes of V2 represent information regarding edge and color features ("what") [Sincich and Horton 2005]. It is logical to suggest that the motion popout and depth discontinuity encoded (predicted) by the thick stripes provide "objectness" context to the thin and pale stripes of the same retinotopic area: for example, area of coherent motion or similar depth is likely to correspond to the same object, so features thereof should be bound together; while depth or motion discontinuity indicates a boundary. Indeed, substantial population of the "border ownership" cells is found in V2. Conversely, color and contour continuity context provided by the "what" prediction may improve predictions in the "where" stream.

**Motor control**: Forward models of motor control are fundamentally predictive in their nature [Miall and Wolpert 1996, Shadmehr et al. 2010]. A forward model predicts the sensory consequences of a motor command. That is, given the internal representation of the current state of the body and its surroundings, and given the current motor command, it predicts the internal representation of the next state of the body and its surroundings. Our cortical model is well suited for implementing the forward model of motor control. Indeed, the top level of the sensory hierarchy may predict the next sensory state when given the motor command (as context) and the current sensory state (as feedforward input from the lower sensory areas).

**Computation**: Our model is well suited to be run on a GPU but we have not had the chance to implement this yet. Further, specialized hardware optimized for convolution or neuromorphic computing are likely good targets. Having access to faster compute resources would allow us to increase the size of these models by several orders of magnitude in order to simulate a model more comparable to the visual acuity of the human visual system.

## Acknowledgments

The authors thankfully acknowledge the support of this research by AFRL and DARPA (grant FA8750-15-C-0178) under the DARPA Cortical Processor Seedling led by Dan Hammerstrom.